\documentclass{article}
\usepackage{hyperref}
\usepackage{datetime}
\usepackage{graphicx}
\usepackage{natbib}
\usepackage{booktabs}
\hypersetup{colorlinks=true, allcolors=blue}

\title{Reevaluating Automated Wildlife Species Detection: A Reproducibility Study on a Custom Image Dataset}

\author{Tobias Abraham Haider}

\newdate{articleDate}{8}{12}{2025}
\date{\displaydate{articleDate}}

\begin{document}
\maketitle
\begin{abstract}This study revisits the findings of Carl et al. \cite{Carl2020}, who evaluated the pre-trained Google Inception-ResNet-v2 model for automated detection of European wild mammal species in camera trap images. To assess the reproducibility and generalizability of their approach, we reimplemented the experiment from scratch using openly available resources and a different dataset consisting of 900 images spanning 90 species. After minimal preprocessing, we obtained an overall classification accuracy of 62\%, closely aligning with the 71\% reported in the original work despite differences in datasets. As in the original study, per-class performance varied substantially, as indicated by a macro F1 score of 0.28, highlighting limitations in generalization when labels do not align directly with ImageNet classes. Our results confirm that pretrained convolutional neural networks can provide a practical baseline for wildlife species identification but also reinforce the need for species-specific adaptation or transfer learning to achieve consistent, high-quality predictions.\end{abstract}\begin{center}\end{center}

\section{Introduction}

While biodiversity is decreasing at a rapid pace, the rise of specific species, be they invasive or predatory, concerns societies around the world. As a consequence, researchers and conservationists are interested in continuously monitoring wildlife populations in terms of their geographical distribution, size, and behavior. Researchers successfully deploy camera traps that can take photographs of passing animals without disturbing them \cite{trolliet2014camera}. The photos are typically manually collected from the traps and annotated with the name of the species present in the image \cite{10.1145/3615893.3628760}.

Deep convolutional neural networks (CNNs) have emerged as a promising solution to automate this process, offering robust image classification capabilities. Building on prior work, this study evaluates the reproducibility of results reported by Carl et al. \cite{Carl2020}, who applied a pre-trained Inception-ResNet-v2 model for European mammal species detection in camera trap images. By reconstructing their experiment using a different dataset and a reproducible, open-source workflow, we examine both the reliability of the original findings and the generalizability of pretrained CNNs to broader wildlife monitoring scenarios.

\section{Experiment Setup}

We reimplemented the Python code for the experiment from scratch because all the necessary components (data \cite{banerjee2024animal}, model \cite{szegedy2016inceptionv4inceptionresnetimpactresidual}, and metrics \cite{scikit-learn}) can be taken from stable public sources. To maximize the readability and reproducibility of the experiment, a minimal setup was chosen, defining all necessary code, data, and requirements in one GitHub project \cite{tobias_haider_2025_17241649}.
State-of-the-art Python packages are chosen, installed, and imported. The exact versions are shown in Table~\ref{table-requirements}. The Jupyter notebook is run locally on a Thinkpad T14 with an AMD Ryzen 5 PRO 5650U processor and 16 GB of memory but no GPU. The operating system is Linux Mint 22.1 and the Python kernel is version 3.12. We expect there to be no deviation in the results, even if different hardware or runtime is chosen.

\begin{table}
\centering
\caption[]{Runtime dependencies

\begin{tabular}{p{\dimexpr 0.500\linewidth-2\tabcolsep}p{\dimexpr 0.500\linewidth-2\tabcolsep}}
\toprule
package & version \\
\hline
pathlib & 1.0.1 \\
Pillow & 11.3.0 \\
numpy & 2.1.3 \\
pandas & 2.3.1 \\
tensorflow & 2.19.0 \\
scikit-learn & 1.7.1 \\
\bottomrule
\end{tabular}}
\label{table-requirements}
\end{table}

\section{Model}

After setting up the Python runtime and importing the required packages, we load the Inception-ResNet-v2 model from the TensorFlow model repository \cite{tensorflow2015-whitepaper}. We use the publicly available pretrained weights, which were obtained by training the model on the ImageNet dataset \cite{5206848}. This approach eliminates the model design and training phase completely but limits the model prediction space to the 1000 classes from the ImageNet dataset.

\begin{verbatim}
model = InceptionResNetV2(weights="imagenet")
\end{verbatim}

\section{Data}

Carl et al. provide the source of the wildlife images used in their dataset \cite{nationalpark2019schwarzwild}. This source is no longer available, requiring us to run the experiment on a different dataset. To test the generalizability of the model, we take a larger public dataset containing images of 90 different species \cite{banerjee2024animal}. To mimic the original experiment setup, only 10 samples are used for each species, resulting in a total test sample size of 900 images.

\includegraphics[width=0.375\linewidth]{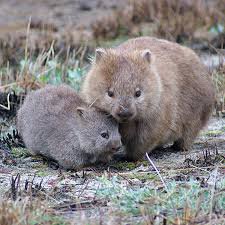}

\subsection{Data Preprocessing}

We load the images, respecting all three color channels (RGB), resize them to 299 by 299 pixels, and convert them into a 1-dimensional vector. The color intensities are scaled to be floating-point numbers from 0 to 1. This is the minimal preprocessing required to fit the required input size of the neural network.

\begin{verbatim}
def load_image(path, target_size):
    image = Image.open(path).convert("RGB")
    image = image.resize(target_size)
    return np.array(image) / 255.0
\end{verbatim}

Then we construct the testing dataset by stacking all normalized image vectors and using the folder names as the labels.

\begin{verbatim}
animal_images = [load_image(p, input_shape) 
                 for p in wildlife_image_paths]
animal_species = [p.parent.name 
                  for p in wildlife_image_paths]

X_test = np.stack(animal_images, axis=0)
y_true = animal_species
\end{verbatim}

\section{Test}

The Inception-ResNet-v2 model outputs a probability distribution over 1,000 classes, corresponding to the categories defined in the ImageNet dataset. For this study, we use only the top-1 prediction (the class with the highest softmax probability) as the model's output and compare it to the ground-truth label from our test dataset.

\begin{verbatim}
y_pred = model.predict(X_test)
y_pred = [pred[0][1] # take output label
          for pred 
          in decode_predictions(y_pred, top=1)]
\end{verbatim}

When looking at the predictions, it becomes apparent that the model yields usable results. Almost all inference outputs are animal species somehow related to the one present in the image. The data already shows that the InceptionResNetV2 is generalizable to some extent.

\begin{table}
\centering
\caption[]{Subset of Inception-ResNet-v2 raw predictions

\begin{tabular}{p{\dimexpr 0.500\linewidth-2\tabcolsep}p{\dimexpr 0.500\linewidth-2\tabcolsep}}
\toprule
y\_pred & y\_true \\
\hline
gazelle & antelope \\
badger & badger \\
hummingbird & bat \\
brown\_bear & bear \\
bee & bee \\
honeycomb & beetle \\
bison & bison \\
wild\_boar & boar \\
ringlet & butterfly \\
Egyptian\_cat & cat \\
\bottomrule
\end{tabular}}
\label{species_recognition_result_raw_short}
\end{table}

\subsection{Label Mapping}

A key challenge in this setup is that ImageNet classes do not align directly with the wildlife species in our dataset. To enable evaluation, we constructed a manual mapping table linking model output labels to the target species. This mapping followed a best-effort approach based on the Linnean system of taxonomy.

Several limitations arise from this process:

\begin{itemize}
\item When we collapse multiple species into higher-level taxa (e.g., mapping all bear species to ``bear''), we lose some species-level detail.
\item Certain species in the dataset are not represented in ImageNet classes at all (e.g., bats, deer), preventing meaningful predictions for these cases.
\end{itemize}

While this mapping introduces ambiguity, it reflects a realistic challenge when applying pretrained ImageNet models to ecological data and illustrates the need for task-specific model adaptation.

\begin{table}
\centering
\caption[]{Mapping rules between ImageNet classes and test data classes

\begin{tabular}{p{\dimexpr 0.500\linewidth-2\tabcolsep}p{\dimexpr 0.500\linewidth-2\tabcolsep}}
\toprule
ImageNet label & dataset label \\
\hline
gazelle, impala & antelope \\
American\_black\_bear, brown\_bear & bear \\
ground\_beetle, leaf\_beetle, rhinoceros\_beetle, dung\_beetle & beetle \\
wild\_boar & boar \\
ringlet, monarch, sulphur\_butterfly, lycaenid & butterfly \\
Egyptian\_cat, tabby, Siamese\_cat, Persian\_cat, lynx & cat \\
water\_buffalo & cow \\
Dungeness\_crab, fiddler\_crab, rock\_crab, king\_crab & crab \\
magpie, jay & crow \\
red\_deer, elk & deer \\
\bottomrule
\end{tabular}}
\label{imagenet_label_mapping}
\end{table}

\section{Evaluation}

Carl et al. reported two performance metrics for their study: overall classification accuracy across the dataset and per-species accuracy. To enable a direct comparison, we adopt the same evaluation strategy. After applying the label mapping described above, predictions are grouped by true species, and accuracy is computed at both the global and class levels. Additionally, we compute the macro F1 score to quantify the per-class imbalance observed in the results.

\begin{table}
\centering
\caption[]{Prediction accuracy per species and the total accuracy

\begin{tabular}{p{\dimexpr 0.500\linewidth-2\tabcolsep}p{\dimexpr 0.500\linewidth-2\tabcolsep}}
\toprule
species & accuracy \\
\hline
bison, bear, boar, crab, elephant, eagle, dog, chimpanzee, cockroach, snake, panda, pelecaniformes, pig, koala, orangutan, ladybug, leopard, lobster, hornbill, jellyfish, hyena, hummingbird, goose, goldfish, fox, fly, sandpiper, zebra, wombat, turtle & 1 \\
parrot, shark, starfish, squirrel, otter, kangaroo, penguin, coyote, butterfly, flamingo, badger, bee, antelope, hare, gorilla, porcupine, tiger, hamster & 0.9 \\
sheep, lizard, lion, cat, dragonfly, wolf & 0.8 \\
beetle, hippopotamus, ox, grasshopper & 0.7 \\
whale & 0.6 \\
duck & 0.4 \\
owl, goat, crow & 0.3 \\
swan & 0.1 \\
caterpillar, bat, dolphin, donkey, cow, deer, mosquito, horse, hedgehog, okapi, moth, mouse, octopus, seal, raccoon, rat, possum, pigeon, oyster, seahorse, rhinoceros, reindeer, squid, sparrow, turkey, woodpecker & 0 \\
TOTAL & 0.62 \\
\bottomrule
\end{tabular}}
\label{species_recognition_result_summary}
\end{table}

\section{Summary}

Our reproduced results confirm the findings of Carl et al. We achieve an overall top-1 prediction accuracy of 62\% by using a larger dataset that includes many species not present in the original study. This result is comparable to the 71\% reported by Carl et al. Consistent with their study, we observe substantial variation in per-species accuracies, signaled by a macro F1 score of 0.28. 48 species out of 90 are predicted with an accuracy greater than or equal to 90\%, and 26 species are predicted without any success (0\% accuracy). A detailed summary of per-species prediction accuracies is provided in Table~\ref{species_recognition_result_summary}.

The experiment shows that pretrained convolutional neural networks, like the Inception-ResNet-v2, are a viable option for the annotation of camera trap images. The network design allows for detailed pattern recognition and robust identification of a large number of animal species. The main issue encountered is the model weights, which are fit for a fixed set of animal species but unsuitable for many classes in our dataset.

\section{Future Work}

Previous research has explored transfer learning of convolutional neural networks for recognizing animal species, showing that retraining pretrained networks for a specific use case can substantially improve prediction accuracy \cite{Islam2023}. We expect this to be a key way to overcome low prediction accuracies for certain species.

We emphasize that, although highly powerful, the models investigated in most studies remain very deep and too large for large-scale deployment in nature. The Inception-ResNet-v2 uses about 55 million parameters, requiring a significant amount of memory and energy. Other models such as MobileNet \cite{9008835} and EfficientNet \cite{tan2020efficientnetrethinkingmodelscaling} use far fewer layers and are optimized for edge deployment.

For future work, we suggest combining these two adaptations (transfer learning and smaller neural networks) and reevaluating this experiment. This effort may yield a highly efficient and accurate model suitable for scalable deployment in real-world wildlife monitoring.

\bibliography{main.bib}

@article{Carl2020,
	author = {Carl, Christin and Sch{\" o}nfeld, Fiona and Profft, Ingolf and Klamm, Alisa and Landgraf, Dirk},
	journal = {European Journal of Wildlife Research},
	doi = {10.1007/s10344-020-01404-y},
	issn = {1439-0574},
	number = {4},
	year = {2020},
	month = {7},
	pages = {62},
	title = {Automated detection of {European} wild mammal species in camera trap images with an existing and pre-trained computer vision model},
	url = {https://doi.org/10.1007/s10344-020-01404-y},
	howpublished = {https://doi.org/10.1007/s10344-020-01404-y},
	volume = {66},
}

@article{trolliet2014camera,
	author = {Trolliet, Franck and Huynen, Marie-Claude and Vermeulen, C{\' e}dric and Hambuckers, Alain},
	journal = {Biology Agriculture Science Environnement},
	year = {2014},
	month = {1},
	pages = {446--454},
	title = {Use of camera traps for wildlife studies. {A} review},
	volume = {18},
}

@inproceedings{10.1145/3615893.3628760,
	address = {Hamburg, Germany},
	author = {Tulasi, Dhruv and Granados, Alys and Gunawardane, Prabath and Kashyap, Abhay and McDonald, Zara and Thulasidasan, Sunil},
	series = {GeoWildLife '23},
	booktitle = {Proceedings of the 1st {ACM} {SIGSPATIAL} {International} {Workshop} on {AI}-{Driven} {Spatio}-{Temporal} {Data} {Analysis} for {Wildlife} {Conservation}},
	doi = {10.1145/3615893.3628760},
	isbn = {9798400703553},
	year = {2023},
	pages = {9--16},
	organization = {Association for Computing Machinery},
	title = {Smart {Camera} {Traps}: Enabling {Energy}-{Efficient} {Edge}-{AI} for {Remote} {Monitoring} of {Wildlife}},
	url = {https://doi.org/10.1145/3615893.3628760},
}

@misc{banerjee2024animal,
	author = {Banerjee, Sourav},
	year = {2024},
	note = {Accessed: 2025-09-03},
	title = {Animal {Image} {Dataset} (90 {Different} {Animals})},
	url = {https://www.kaggle.com/datasets/iamsouravbanerjee/animal-image-dataset-90-different-animals},
	howpublished = {https://www.kaggle.com/datasets/iamsouravbanerjee/animal-image-dataset-90-different-animals},
}

@misc{szegedy2016inceptionv4inceptionresnetimpactresidual,
	author = {Szegedy, Christian and Ioffe, Sergey and Vanhoucke, Vincent and Alemi, Alex},
	year = {2016},
	title = {Inception-v4, {Inception}-{ResNet} and the {Impact} of {Residual} {Connections} on {Learning}},
	url = {https://arxiv.org/abs/1602.07261},
	howpublished = {https://arxiv.org/abs/1602.07261},
}

@article{scikit-learn,
	author = {Pedregosa, F. and Varoquaux, G. and Gramfort, A. and Michel, V. and Thirion, B. and Grisel, O. and Blondel, M. and Prettenhofer, P. and Weiss, R. and Dubourg, V. and Vanderplas, J. and Passos, A. and Cournapeau, D. and Brucher, M. and Perrot, M. and Duchesnay, E.},
	journal = {Journal of Machine Learning Research},
	year = {2011},
	pages = {2825--2830},
	title = {Scikit-learn: Machine {Learning} in {Python}},
	volume = {12},
}

@misc{tobias_haider_2025_17241649,
	author = {Haider, Tobias},
	doi = {10.5281/zenodo.17241649},
	year = {2025},
	month = {10},
	publisher = {Zenodo},
	title = {tobsel7/research-vetmedwien-animal-species- identification: More detailed evaluation and description of experiment results},
	url = {https://doi.org/10.5281/zenodo.17241649},
}

@misc{tensorflow2015-whitepaper,
	author = {Abadi, Mart{\' i}n and Agarwal, Ashish and Barham, Paul and Brevdo, Eugene and Chen, Zhifeng and Citro, Craig and Corrado, Greg S. and Davis, Andy and Dean, Jeffrey and Devin, Matthieu and Ghemawat, Sanjay and Goodfellow, Ian and Harp, Andrew and Irving, Geoffrey and Isard, Michael and Jia, Yangqing and Jozefowicz, Rafal and Kaiser, Lukasz and Kudlur, Manjunath and Levenberg, Josh and Man{\' e}, Dandelion and Monga, Rajat and Moore, Sherry and Murray, Derek and Olah, Chris and Schuster, Mike and Shlens, Jonathon and Steiner, Benoit and Sutskever, Ilya and Talwar, Kunal and Tucker, Paul and Vanhoucke, Vincent and Vasudevan, Vijay and Vi{\' e}gas, Fernanda and Vinyals, Oriol and Warden, Pete and Wattenberg, Martin and Wicke, Martin and Yu, Yuan and Zheng, Xiaoqiang},
	year = {2015},
	note = {Software available from tensorflow.org},
	title = {TensorFlow: Large-{Scale} {Machine} {Learning} on {Heterogeneous} {Systems}},
	url = {https://www.tensorflow.org/},
	howpublished = {https://www.tensorflow.org/},
}

@inproceedings{5206848,
	author = {Deng, Jia and Dong, Wei and Socher, Richard and Li, Li-Jia and Li, Kai and Fei-Fei, Li},
	booktitle = {2009 {IEEE} {Conference} on {Computer} {Vision} and {Pattern} {Recognition}},
	doi = {10.1109/CVPR.2009.5206848},
	year = {2009},
	pages = {248--255},
	title = {ImageNet: A large-scale hierarchical image database},
}

@misc{nationalpark2019schwarzwild,
	author = {{Nationalparkverwaltung Hainich, FFK Gotha}},
	year = {2019},
	title = {Schwarzwildforschung im {Hainich}},
	url = {https://www.schwarzwild-hainich.de},
	howpublished = {https://www.schwarzwild-hainich.de},
}

@article{Islam2023,
	author = {Binta Islam, Sazida and Valles, Damian and Hibbitts, Toby and Ryberg, Wade and Walkup, Danielle and Forstner, M.},
	journal = {Animals},
	doi = {10.3390/ani13091526},
	year = {2023},
	month = {5},
	pages = {1526},
	title = {Animal {Species} {Recognition} with {Deep} {Convolutional} {Neural} {Networks} from {Ecological} {Camera} {Trap} {Images}},
	volume = {13},
}

@inproceedings{9008835,
	author = {Howard, Andrew and Sandler, Mark and Chen, Bo and Wang, Weijun and Chen, Liang-Chieh and Tan, Mingxing and Chu, Grace and Vasudevan, Vijay and Zhu, Yukun and Pang, Ruoming and Adam, Hartwig and Le, Quoc},
	booktitle = {2019 {IEEE}/{CVF} {International} {Conference} on {Computer} {Vision} ({ICCV})},
	doi = {10.1109/ICCV.2019.00140},
	year = {2019},
	pages = {1314--1324},
	title = {Searching for {MobileNetV3}},
}

@misc{tan2020efficientnetrethinkingmodelscaling,
	author = {Tan, Mingxing and Le, Quoc V.},
	year = {2020},
	title = {EfficientNet: Rethinking {Model} {Scaling} for {Convolutional} {Neural} {Networks}},
	url = {https://arxiv.org/abs/1905.11946},
	howpublished = {https://arxiv.org/abs/1905.11946},
}
\end{document}